%% file: cvpr.tex
\documentclass[final]{cvpr}
\input{math_commands.tex}

\usepackage{url}
\usepackage[utf8]{inputenc} % allow utf-8 input
\usepackage[T1]{fontenc}    % use 8-bit T1 fonts
\usepackage{url}            % simple URL typesetting
\usepackage{booktabs}       % professional-quality tables
\usepackage{amsfonts}       % blackboard math symbols
\usepackage{nicefrac}       % compact symbols for 1/2, etc.
\usepackage{microtype}      % microtypography

\usepackage{amsmath}
\usepackage{amssymb}
\usepackage{wrapfig}
\usepackage{subcaption}
\usepackage{multirow}
 \usepackage{mathtools} 
\usepackage{verbatim}
\usepackage{anyfontsize}
\usepackage{color,xcolor}
\newcommand{\distas}[1]{\mathbin{\overset{#1}{\kern\z@\sim}}}%
\usepackage{enumitem}
\usepackage[lined,boxed,commentsnumbered,ruled,linesnumbered]{algorithm2e}
\usepackage{colortbl} % rowcolor
\definecolor{Gray}{gray}{0.93}
\usepackage{CJK}

\include{mlVecMat}

\usepackage{times}
\usepackage{epsfig}
\usepackage{graphicx}
\usepackage{amsmath}
\usepackage{amssymb}
\usepackage[pagebackref=true,breaklinks=true,colorlinks,bookmarks=false]{hyperref}

\def\cvprPaperID{8012} % *** Enter the CVPR Paper ID here
\def\confYear{CVPR 2021}

\begin{document}

%%%%%%%%% TITLE
\title{M$^3$P: Learning Universal Representations via Multitask Multilingual Multimodal Pre-training}

\author{Minheng Ni$^1$\footnotemark[1]\ \,\footnotemark[2] \quad Haoyang Huang$^2$\footnotemark[2] \quad Lin Su$^3$\footnotemark[2] \quad Edward Cui$^3$ \quad Taroon Bharti$^3$ \quad Lijuan Wang$^4$\\
Jianfeng Gao$^5$ \quad Dongdong Zhang$^2$ \quad Nan Duan$^2$\footnotemark[3]\\
$^1$ Research Center for Social Computing and Information Retrieval\\
Harbin Institute of Technology, China\\
$^2$ Natural Language Computing, Microsoft Research Asia, China\\
$^3$ Bing Multimedia Team, Microsoft, China\\
$^4$ Cloud+AI, Microsoft, United States \\
$^5$ Deep Learning, Microsoft Research Redmond, United States \\
{\tt\small mhni@ir.hit.edu.cn} \\
{\tt\small \{haohua, lins, edwac, tbharti, lijuanw, jfgao,  Dongdong.Zhang, nanduan\}@microsoft.com}
}

\maketitle
\renewcommand{\thefootnote}{\fnsymbol{footnote}}

\footnotetext[1]{Work is done during an internship at Microsoft Research Asia.}
\footnotetext[2]{These authors contributed equally to this work.}
\footnotetext[3]{Corresponding Author.}

\newcommand{\fix}{\marginpar{FIX}}
\newcommand{\new}{\marginpar{NEW}}

\newcommand{\modelshort}{M$^3$P}

\newcommand{\taskone}   {Multilingual Masked Language Modeling}
\newcommand{\tasktwo}   {Multilingual Denoising Auto-Encoding}
\newcommand{\taskthree} {Multimodal Masked Language Modeling}
\newcommand{\taskfour}  {Denoising Image Captioning}
\newcommand{\taskfive}  {Masked Region Modeling}
\newcommand{\tasksix}   {Visual-Linguistic Matching}
\newcommand{\taskseven} {Image Captioning}

\newcommand{\taskoneshort}  {xMLM}
\newcommand{\tasktwoshort}  {xDAE}
\newcommand{\taskthreeshort}{MMLM}
\newcommand{\taskfourshort} {DIC}
\newcommand{\taskfiveshort} {MRM}
\newcommand{\tasksixshort}  {VLM}
\newcommand{\tasksevenshort}{IC}

%%%%%%%%% BODY TEXT
\input{00_abstract}
\input{01_intro}
\input{02_related_work}
\input{03_method}
\input{04_exps}
\input{05_results}
\input{06_conclusion}

{\small
\bibliographystyle{ieee_fullname}
\bibliography{egbib}
}

\end{document}

%% file: math_commands.tex
\usepackage{amsmath,amsfonts,bm}

\def\eqref#1{equation~\ref{#1}}
\def\1{\bm{1}}

\DeclareMathAlphabet{\mathsfit}{\encodingdefault}{\sfdefault}{m}{sl}
\SetMathAlphabet{\mathsfit}{bold}{\encodingdefault}{\sfdefault}{bx}{n}

%% file: mlVecMat.tex
\newcommand{\RN}[1]{%
	\textup{\lowercase\expandafter{\it \romannumeral#1}}%
}

\newcommand{\beq}{\vspace{0mm}\begin{equation}}
\newcommand{\eeq}{\vspace{0mm}\end{equation}}
\newcommand{\beqs}{\vspace{0mm}\begin{eqnarray}}
\newcommand{\eeqs}{\vspace{0mm}\end{eqnarray}}
\newcommand{\barr}{\begin{array}}
\newcommand{\earr}{\end{array}}

\newcommand{\Lcal}{\mathcal{L}}

\DeclareMathOperator{\EE}{\mathbb{E}} % Expectation

%% file: 00_abstract.tex
\begin{abstract}
%We present \textbf{\modelshort}, a \textbf{M}ultitask \textbf{M}ultilingual \textbf{M}ultimodal \textbf{P}re-trained model that combines multilingual pre-training and multimodal pre-training into a unified framework via multitask pre-training. The model is pre-trained with Code-switching to enhance the performance of low-resource language and resolve the tendency among multilingual data and multimodal data. The model learns universal representations that can map objects that occurred in different modalities or expressed in different languages to vectors in a common semantic space. Evaluations show that \modelshort\ can (i) obtain new state-of-the-art results on non-English Image-Text Retrieval task and comparable results on English, and (ii) significantly improve the performance of non-English Image-Text Retrieval task under low-resource settings.

We present \textbf{\modelshort}, a \textbf{M}ultitask \textbf{M}ultilingual \textbf{M}ultimodal \textbf{P}re-trained model that combines multilingual pre-training and multimodal pre-training into a unified framework via multitask pre-training. 
% To enhance generalization ability on low-resource languages, we further combine Multimodal Code-switched Training (MCT) with pre-training tasks.
%Our goal is to learn universal representations that can map objects occurred in different modalities and texts expressed in different languages to vectors in a common semantic space.
Our goal is to learn universal representations that can map objects occurred in different modalities or texts expressed in different languages into a common semantic space.
% In addition to \modelshort\ for non-English image-text alignments, we also propose
% Multimodal Code-switched Training (MCT) via combining multimodal pre-training with code-switch strategy to explicitly encourage fine-grained alignment between images and non-English languages.
In addition, to explicitly encourage fine-grained alignment between images and non-English languages, we also propose Multimodal Code-switched Training (MCT) to combine monolingual pre-training and multimodal pre-training via a code-switch strategy.
% To alleviate lacking enough non-English labeled data, we adopt a Multimodal Code-switched Training (MCT) strategy, and combine it with mutlimodal pre-training to
% explicitly encourage fine-grained alignment between images and non-English languages.
% , and (ii) to alleviate lacking enough non-English labeled data.
% In addition to  alleviate lacking enough labeled data for non-English multimodal tasks, 
%  Specifically, the proposed \modelshort\ also employs a new novel method, Multimodal Code-switched Training (MCT) to explicitly encourage fine-grained alignment between images and non-English languages during pre-training.
% we also propose Multimodal Code-switched Training (MCT) to explicitly encourage fine-grained alignment between images and non-English languages during pre-training.
%A code-switching strategy is further proposed to enhance the generalization ability of \textbf{\modelshort} on low-resource languages. 
%To alleviate lacking enough labeled data for non-English image-language tasks, we adopt Multimodal Code-switched Training (MCT) to enhance the generalization ability of \modelshort\ on low-resource languages.
%We adopt Multimodal Code-switched Training (MCT) to alleviate lacking enough labeled data for low-resource languages.
%Evaluation results show that \modelshort\ can obtain state-of-the-art results on MSCOCO and Multi30K for non-English image-text retrieval tasks.
Experiments are performed on the multilingual image retrieval task across two benchmark datasets, including MSCOCO and Multi30K. \modelshort\ can achieve comparable results for English and new state-of-the-art results for non-English languages.

% Comprehensive analysis shows that both conditional masking and OT-based WRA contribute to better pre-training.

\end{abstract}

%% file: 01_intro.tex
\section{Introduction}
\label{sec:intro}

Recently, we witness the rise of a new paradigm of natural language processing (NLP), where general knowledge is learned from raw texts by self-supervised pre-training and then applied to downstream tasks by task-specific fine-tuning. Now, these state-of-the-art monolingual pre-trained language models, such as BERT \cite{devlin2019bert}, RoBERTa \cite{liu2019roberta} and GPT-2 \cite{radford2019language}, have been expanded to \textit{multilingual scenarios}, such as Multilingual BERT \cite{devlin2019bert}, XLM/XLM-R \cite{conneau2019xlm, conneau2019unsupervised}, Unicoder \cite{huang2019unicoder}. Moreover, some pre-training models under \textit{multimodal scenarios}, such as Unicoder-VL \cite{li2019unicodervl}, UNITER \cite{chen2019uniter}, ERNIE-ViL \cite{yu2020ernie}, VILLA \cite{gan2020large} and Oscar \cite{li2020oscar}, also come out.

However, it is still challenging to extend these pre-trained models to multilingual-multimodal scenarios. The multilingual pre-trained language models cannot handle vision data (e.g., images or videos) directly, whereas many pre-trained multimodal models are trained on English corpora thus cannot perform very well on non-English languages. Therefore, high quality multilingual multimodal training corpus is essential to combine multilingual pre-training and multimodal pre-training. However, there are only a few multilingual multimodal corpora exist, and they also have low language coverage.
%Due to the lack of language coverage in existing multimodal corpora (which are mostly in English only),
%many multilingual pre-trained models cannot handle vision data (e.g., images and videos), whereas many multimodal pre-trained models, which are trained using texts mainly in English, cannot handle multiple languages. 
Moreover, relying on high-quality machine translation engines to generate such data from English multimodal corpora is both time-consuming and computationally expensive. Learning explicit alignments between vision and non-English languages during pre-training is lacking. %Efficient strategy to model the explicit alignments between vision and non-English languages during pre-training also impedes research in the multilingual multimodal space.

To address these challenges, this paper presents \textbf{\modelshort}, a \textbf{M}ultitask \textbf{M}ultilingual \textbf{M}ultimodal \textbf{P}re-trained model, which aims to learn universal representations that can map objects occurred in different modalities or texts expressed in different languages into a common semantic space. In order to alleviate the issue of lacking enough non-English labeled data for multimodal pre-training, we introduce \textit{Multimodal Code-switched Training} (MCT) to enforce the explicit alignments between images and non-English languages. The goal is achieved by (i) learning to represent multilingual data using multilingual corpora (e.g., sentences from Wikipedia covering 100 languages) by multilingual pre-training, (ii) learning multilingual-multimodal representations by randomly replacing some English words with their translations in other languages from multimodal corpora (e.g., image-caption pairs labeled in English), and (iii) generalizing these representations to deal with multilingual-multimodal tasks by Multitask learning.

In summary, the main contributions of this paper are:

\begin{itemize}
\item We present \modelshort, the first known effort on combining multilingual pre-training and multimodal pre-training into a unified framework.
% and supports 100 languages for \textit{multilingual-multimodal scenarios}. 

\item We propose a novel \textit{Multimodal Code-switched Training} (MCT) method, an effective way to enhance the multilingual transfer ability of \modelshort\ in the zero-shot and few-shot settings.

% multimodal scenarios

% We propose \textit{Multimodal Code-switched Training}, which is a effective way to catch more explicit alignments contained in
% non-English languages and images regions. 
% In particular, Mutlimodal Code-switched Training 
% improves  the multilingual transfer ability of \modelshort\ in the zero-shot or few-shot setting.

\item We achieve new state-of-the-art results for the multilingual image-text retrieval task on both Multi30K and MSCOCO for non-English languages, outperforming existing multilingual methods by a large margin. The proposed model can also achieve comparable results for English on these two datasets, compared to the state-of-the-art monolingual multimodal models.

% We also present extensive experiments
% and analysis to provide useful insights on the effectiveness of each pre-training task and model
% variant.
% we conduct extensive experiments on Multilingual Image-text Retrieval task across two datasets MSCOCO and Multi30K. \modelshort\ achieves comparable results on English, compared to the coherent English multimodal pre-trained model, and obtains new state-of-the-art results for non-English languages. 

\item  Last but not least, we conduct extensive experiments and analysis to provide insights on the effectiveness of using \textit{Multimodal Code-switched Training} (MCT) and each pre-training task.

% variant

% In summary, we mainly
% make the following contributions:

% is and integrate it into M3P with significant gains on non-English languages in the zero-shot setting or few-shot setting.
% %to further alleviate the issue of lacking enough labeled data for non-English multimodal tasks and enhance the generalization ability of \modelshort\ on low-resource languages.
% \item Experiments on Multilingual Image-text Retrieval dataset MSCOCO and Multi30K show that \modelshort\ achieves comparable results on English, compared to the state-of-the-art pre-trained models. We also obtains new state-of-the-art results for non-English languages and outperforms existing methods by a large margin. 
\end{itemize}

%% file: 02_related_work.tex
\section{Related Work}
\label{sec:related}

\paragraph{Multilingual Pre-trained Models} 
Multilingual BERT (M-BERT) \cite{devlin2019bert} demonstrates that by performing masked language modeling on a multilingual corpus with shared vocabulary and weights for 102 languages, surprisingly good results can be achieved on the cross-lingual natural language inference (XNLI) \cite{conneau2018xnli} task in 15 languages. 
XLM \cite{conneau2019xlm} and Unicoder \cite{huang2019unicoder} further improve the multilingual BERT by introducing new pre-training tasks based on a bilingual corpus.
However, all such models work for NLP tasks and are not well designed for multimodal tasks such as Multilingual Image-text Retrieval or Multimodal Machine Translation.

\paragraph{Multimodal Pre-trained Models}  
Recently, a large number of multimodal pre-trained models, such as ViLBERT \cite{lu2019vilbert}, Unicoder-VL \cite{li2019unicodervl}, UNITER \cite{chen2019uniter}, VLP \cite{zhou2019unified} and Oscar \cite{li2020oscar}, are developed for vision-language tasks using multi-layer Transformer as the backbone.
%These models are pre-trained using similar visual-linguistic tasks and achieve comparable results on many vision-language tasks, such as image-text retrieval.
However, as it is not easy to collect well-aligned visual-linguistic training data in multiple languages, all these models are pre-trained for English only based on monolingual multimodal corpora, such as Conceptual Captions \cite{sharma2018conceptual}, SBU Captions \cite{ordonez2011im2text}, Visual Genome \cite{krishna2017visual} and MSCOCO \cite{chen2015microsoft}. Hence, it is not feasible to apply them into multimodal tasks with non-English inputs.

\paragraph{Code-switched Training}
Code-switched training \cite{qin2020cosda} \cite{yang2020csp} converts the original training corpus to code-switched corpus, which can help the model explicitly model the relationship among corresponding words in different languages. Existing work uses a rule-based word replacement strategy to replace the original word with translated word randomly by bilingual dictionaries. This approach provides a significant improvement to the low-resource language. However, existing works use Code-switching for text-only tasks and ignore its application on multimodal pre-training model under \textit{multilingual-multimodal scenarios}.

%% file: 03_method.tex
\section{M$^3$P: Multitask Multilingual Multimodal Pre-training}

\begin{figure*} [t]
	\centering
	\includegraphics[width=17.5cm]{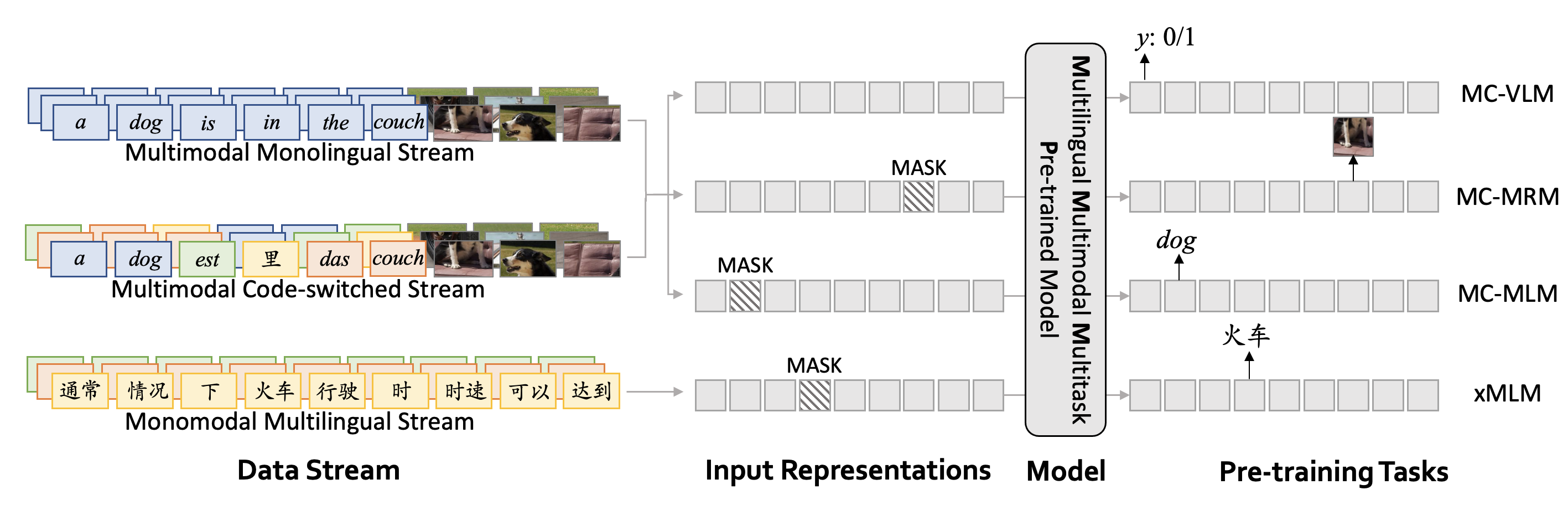}
	\caption{Three data streams and four pre-training tasks used in \modelshort.
		\textcolor[RGB]{55,81,138}{Blue} blocks denote English text, and
		\textcolor[RGB]{246,195,66}{Yellow}, \textcolor[RGB]{126,172,85}{Green} and \textcolor[RGB]{223,131,68}{Orange} blocks denote non-English text.}
	\label{overveiw}
\end{figure*}

In this section, we describe how we train \modelshort\ using a multilingual-monomodal corpus (e.g., sentences extracted from Wikipedia) and a monolingual-multimodal corpus (e.g., English image-caption pairs). As outlined in Figure \ref{overveiw}, we use the self-attentive transformer architecture of BERT, and design two pre-training objectives with three types of data streams. Multitask training is employed into the pre-training stage to optimize all pre-training objectives simultaneously for better performance. We optimize the accumulated loss of both pre-training objectives with the same weight in each iteration to train them by turns.
% Pre-training the model directly on these corpus makes the multilingual training and multimodal training individually.

\subsection{Data Stream}

%The \modelshort\ is pre-trained via multilingual corpus and multimodal corpus. No images are in multilingual corpus whereas all texts are English only in multimodal corpus. 
We use two basic data streams, Multilingual Monomodal Stream and Monolingual Multimodal Stream, from the multilingual corpus and multimodal corpus, respectively. We also design Multimodal Code-switched Stream to utilize multilingual data and multimodal data at the same time.
%The pre-training data of the \modelshort\ is either the single text or the pair of image and text. We convert both the text and image to the representation sequence. Given a pair of image and text, \modelshort\ tasks use its concatenated text and image representation sequence as the input. While given a single text, text representation sequence will be used as a standalone input.
Details regarding the three data streams are introduced below.

\paragraph{Multilingual Monomodal Stream}

To apply multilingual pre-training, we use Multilingual Monomodal Stream as model input. Given an input text in any language $w^{\left[l_i\right]}$, we first tokenize it into a sequence of BPE tokens via Sentence Piece \cite{kudo2018sentencepiece}. Then we can obtain a text representation sequence by summing up the text embedding and the position embedding of each BPE token.  Moreover, a language embedding \cite{conneau2019xlm} is added to each token to indicate its language attribute. Specifically, the input data is defined as:
\begin{equation}
    \left\{\textbf{w}^{\left[l_i\right]}\right\}=\left\{(\textbf{w}_1^{\left[l_i\right]}, \textbf{w}_2^{\left[l_i\right]},...,\textbf{w}_M^{\left[l_i\right]})\right\}  \nonumber
\end{equation}
where $M$ denotes the length of $w^{\left[l_i\right]}$ and $l_i$ denotes a language in the language set $L$. We denote this stream as $D^{\left[\rm X\right]}$.

\paragraph{Monolingual Multimodal Stream}

To apply multimodal pre-training, we use Monolingual Multimodal Stream as model input. Given a pair of English text and image $(w^{\left[\rm EN\right]}, v)$, the text representation sequence of $w^{\left[\rm EN\right]}$ is obtained similarly as we described in Multilingual Monomodal Stream section, where English is used as the language embedding. 
%For the image $v$, we obtain its image representation sequence. The region representation sequence is the sequence outputted by Faster-RCNN\cite{Detectron2018} on an image. 
For the image $v$, we use Faster-RCNN \cite{Detectron2018} to detect image regions and use corresponding visual features in each region as a visual feature sequence. We also add a spatial embedding to each visual token, which is a 5-D vector based on its normalized top-left, bottom-right coordinates, and the fraction of the image area covered. We project these two vectors to the same dimension of the text representation using two fully-connected (FC) layers. Therefore, the image representation sequence is obtained by summing up its projected visual feature vector and spatial embedding vector of each region in the image.
Furthermore, 
%to indicate whether the input is image or text, 
we add a stream tag [IMG] at the beginning of the image region sequence to separate text tokens and image tokens, and concatenate them to form an input stream:
\begin{equation}
\begin{aligned}
    \left\{\textbf{w}^{\left[{\rm EN}\right]}, \textbf{v}\right\}=\left\{(\textbf{w}_1^{\left[{\rm EN}\right]}, \textbf{w}_2^{\left[{\rm EN}\right]},...,\textbf{w}_M^{\left[{\rm EN}\right]}), ({\textbf{v}_1,\textbf{v}_2,...,\textbf{v}_N})\right\} \nonumber
\end{aligned}
\end{equation}
We denote this stream as $D^{\left[\rm EN\right]}$.

\paragraph{Multimodal Code-switched Stream}

%Given a pair of English text and image $(w^{\left[\rm EN\right]}, v)$ and the set of code-switched languages set $ \mathbf{C} = \{c_1, c_2, ..., c_k\}$, we need bilingual dictionaries which can translate word from English to each language $ c_i $.
%Following \cite{qin2020cosda}, for each word in English text $ w^{\left[\rm EN\right]} $, we will replace it with $\beta $ possibility and make it unchanged with $ 1 - \beta $ possibility. For each selected to be replaced word $ w_i $ , we will translate this word into a random language.
%If a word has multiple translations, we choose a random one. We will obtain the text representation sequence of the Code-switched text $w^{\left[\rm C\right]}$. We keep the original language embedding.\footnote{We have tried to change language embedding during Code-switched, but no significant gain was obtained.}
%Similar with Multimodal Monolingual Stream, text and image representation sequences are concatenated as a whole input stream:

We generate Multimodal Code-switched Stream from Monolingual Multimodal Stream by code-switched method, given English text and image pairs $(w^{\left[\rm EN\right]}, v)$, the set of code-switched languages $ \mathbf{C} = \{c_1, c_2, ..., c_k\}$, and bilingual dictionaries which can translate a word from English to any language $ c_i $. Following \cite{qin2020cosda}, for each word $ w_i^{\left[\rm EN\right]} $ in English text $ w^{\left[\rm EN\right]} $, we replace it with a translated word with a probability of $\beta $. 
If a word has multiple translations, we choose a random one. Similar to the generation process of Multilingual Monolingual Stream, we obtain the text representation sequence of the Code-switched text $w^{\left[\rm C\right]}$ in the same way while keeping the original language embedding.\footnote{We have tried to change language embedding in Code-switched Stream, but no significant gain was obtained.}
Similar with Monolingual Multimodal Stream, the text and image representation sequences are concatenated as the final input stream:
$
    \left\{(\textbf{w}_1^{\left[d_1\right]}, \textbf{w}_2^{\left[d_2\right]},...,\textbf{w}_M^{\left[d_M\right]}), ({\textbf{v}_1,\textbf{v}_2,...,\textbf{v}_N})\right\}
$
, where $d_i$ is a random language in $\{{\rm EN}\} \cup \mathbf{C}$. We simplify the input sequence as:
\begin{equation}
\begin{aligned}
    \left\{\textbf{w}^{\left[{\rm C}\right]}, \textbf{v}\right\}=\left\{(\textbf{w}_1^{\left[{\rm C}\right]}, \textbf{w}_2^{\left[{\rm C}\right]},...,\textbf{w}_M^{\left[{\rm C}\right]}), ({\textbf{v}_1,\textbf{v}_2,...,\textbf{v}_N})\right\} \nonumber
\end{aligned}
\end{equation}
We denote this stream as $D^{\left[\rm C\right]}$.

\subsection{Pre-training Objectives}

%Pre-training the model directly on these corpus makes the multilingual training and multimodal training individually. The model will be hard to learn different languages from the shared vision modal and the alignment between vision and non-English texts. 
To pre-train \modelshort\ under \textit{multilingual-multimodal scenario}, we designed two types of pre-training objectives. \textit{Multilingual Training} aims to learn grammar or syntax from well-formed multilingual sentences.
\textit{Multimodal Code-switched Training} (MCT) aims to learn different languages from the shared vision modal and the alignment between vision and non-English texts.

\subsubsection{Multilingual Training}

\paragraph{\taskone\ (\taskoneshort)}
Similar to Multilingual BERT \cite{devlin2019bert}, XLM \cite{conneau2019xlm} and Unicoder \cite{huang2019unicoder}, this task performs masked language modeling based on the multilingual corpus. At each iteration, a batch is composed of sentences sampled from different languages. The sampling probability of a language $l_i$ is defined as $\lambda_{l_i}= p_{l_i}^{\alpha} / \sum_{l_i} p_{l_i}^{\alpha}$,
where $p_{l_i}$ is the percentage of $l_i$ in the entire multilingual corpus, and the smoothing factor $\alpha$ is set to 0.3.
In each batch, we randomly sample 15\% of the words and (i) replace them with a special symbol [MASK], (ii) replace them with random tokens, or (iii) keep them unchanged with a probability of 80\%, 10\% and 10\%, respectively. We only use Multilingual Monomodal Stream $D^{\left[\rm X\right]}$ for we do not need to use Code-switching to extend it to multilingual corpus. The loss function is defined as:
\begin{equation}
	\begin{aligned}
		\Lcal_{\rm xMLM}(\theta)=
		-\EE_{\textbf{w}^{\left[l_i \right]}\sim D^{\left[\rm X\right]}} \log q_\theta (w_m^{\left[l_i \right]} |\textbf{w}^{\left[l_i \right]}_{\backslash m}) \nonumber
	\end{aligned}
\end{equation}
, where $w^{\left[l_i\right]}_m$ is the masked token and $\textbf{w}^{\left[l_i \right]}_{\backslash m}$ is its context.

\subsubsection{Multimodal Code-switched Training}

Because of the lack of labeled data for the non-English multimodal scenario, the model can only learn multilingualism and multimodality independently. To help the model learn different language representations under the shared vision modal, we propose three Multimodal Code-switched Training tasks: MC-MLM, MC-MRM and MC-VLM. We mix Multimodal Code-switched Stream $D^{\left[\rm C\right]}$ and Monolingual Multimodal Stream $D^{\left[\rm EN\right]}$ with a proportion ratio of $\alpha$ and $1 - \alpha$, respectively, in train these tasks. To simplify the symbols, we denote the mixed data stream as $D$ and omit the mask $\left[ {\rm EN}\right]$ or $\left[ {\rm C}\right]$ as $\left[ {\rm \cdot}\right]$. %for the representation sequence in the mixed data stream.

\paragraph{Multimodal Code-switched Masked Language Modeling (MC-MLM)}
Different from the pre-training tasks in ViLBERT \cite{lu2019vilbert} and Unicoder-VL \cite{li2019unicodervl}, this task aims to learn the representation of different languages based on the shared vision modal. Mixed data stream $D$ is used for training this objective. Specifically, the model predicts each masked token $w^{\left[ {\rm \cdot}\right]}_m$ in the caption $\textbf{w}^{\left[ {\rm \cdot}\right]}$ based on its surrounding tokens $\textbf{w}^{\left[ {\rm \cdot}\right]}_{\backslash m}$ and all image regions $\textbf{v}$. We follow the same masking strategy used in \taskoneshort\ to mask tokens in the input caption. The loss function is defined as:
\begin{equation}
	\begin{aligned}
		\Lcal_{\rm MC-MLM}(\theta)=
		-\EE_{(\textbf{w}^{\left[ {\rm \cdot}\right]},\textbf{v})\sim D} \log p_\theta (w^{\left[ {\rm \cdot}\right]}_m |\textbf{w}^{\left[ {\rm \cdot}\right]}_{\backslash m},\textbf{v}) \nonumber
	\end{aligned}
\end{equation}
, where $D$ is the mixed data stream.

\paragraph{Multimodal Code-switched Masked Region Modeling (MC-MRM)}
This task aims to learn vision representations with multilingual text as the context in mixed data stream $D$.
The model reconstructs each masked image region $v_n$ based on the remaining regions $\textbf{v}_{\backslash n}$ and all caption tokens $\textbf{w}^{\left[ {\rm \cdot}\right]}$.
We randomly mask image regions with a probability of 15\%. The input representation of each masked image region is set to zeros or kept as the original values with a probability of 90\% and 10\%, respectively. 
We apply an FC layer to convert the Transformer output of each masked region $v_k$ into a vector $h_\theta(v_k)$ of the same dimension with the visual feature $f(v_k)$.
We use cross-entropy loss ${\rm CE}(g_\theta(v_k),C(v_k))$ to predict the object category of each masked region $v_k$.
We also apply another FC layer to convert the Transformer output of each masked region $v_k$ to predict the scores of $K$ object classes, which further go through a softmax function to be transformed into a normalized distribution $g_\theta(v_k)$.
We take the predicted object category with the highest confidence score outputted by Faster-RCNN as the ground-truth label of $v_k$, and convert it into a one-hot vector $C(v_k) \in \mathbb{R}^K$. 
Due to the top-1 category predicted by Faster-RCNN is not always correct, we leave minimizing the KL divergence between two distributions for our future work. The loss function can be defined as:
\begin{equation}
	\begin{aligned}
		\Lcal_{\rm MC-MRM}(\theta) = -\EE_{(\textbf{w}^{\left[ {\rm \cdot}\right]},\textbf{v})\sim D}\sum_{k}[&{\rm MSE}(h_\theta(v_k),f(v_k)) + \\
		&{\rm CE}(g_\theta(v_k),C(v_k))] \nonumber
	\end{aligned}
\end{equation}
where $k$ enumerates the index of each masked image region and ${\rm MSE}(h_\theta(v_k),f(v_k))$ denotes the mean-square-error loss that regresses the Transformer output of each masked region $v_k$ to its visual feature $f(v_k)$.

\paragraph{Multimodal Code-switched Visual-Linguistic Matching (MC-VLM)}
\label{sec:VLM}
This task aims to learn alignment between multilingual texts and images with mixed data stream $D$. An FC layer $s_\theta(\textbf{w}^{\rm \left[\cdot\right]},\textbf{v})$ is applied on the Transformer output of [CLS] to predict whether the input image $\textbf{v}$ and the input English or Code-switched text $\textbf{w}^{\rm \left[\cdot \right]}$ are semantically matched. Negative image-caption pairs are created by replacing the image or text in a matched sample with a randomly-selected image or text from other samples. We use Binary Cross-Entropy as the loss function:
\begin{equation}
	\begin{aligned}
		\Lcal_{\rm MC-VLM}(\theta)=-E_{(\textbf{w}^{\left[ {\rm \cdot}\right]},\textbf{v})\sim D}[{\rm BCE}(s_{\theta}(\textbf{w}^{\left[ {\rm \cdot}\right]}, \textbf{v}), y)] \nonumber
	\end{aligned}
\end{equation}
where $y \in \{{0,1}\}$ indicates whether the input image-text pair is matched and ${\rm BCE}$ indicates binary-cross-entropy loss.

%% file: 04_exps.tex
\section{Experiments}

In this section, we describe detailed experimental settings during pre-training, fine-tuning and evaluating \modelshort\ model.
%, and its evaluation results on multilingual image-text retrieval task.

\subsection{Dataset Description}

\begin{table}[t]
\small
\centering
\begin{tabular}{l|ccc}
\toprule
Dataset & Images & Texts & Langauges  \\
\midrule
\midrule
\multicolumn{4}{l}{\textit{Pre-training Corpus}}\\
\midrule
Wikipedia & - & 101G & 100 \\
Conceptual Captions \cite{sharma2018conceptual} & 3.3M & 3.3M & 1 \\
\midrule
\midrule
\multicolumn{4}{l}{\textit{Fine-tuning and Evaluation Corpus}}\\
\midrule
Multi30K \cite{young2014flick}            & 32K  & 384K & 5 \\
MSCOCO \cite{chen2015microsoft} \cite{yoshikawa2017stair} \cite{mcoco2}            & 120K & 1.5M & 3 \\
\bottomrule
\end{tabular}
\caption{\label{dataset}Statistics of datasets.}
\end{table}

As shown in Table \ref{dataset}, we construct our pre-training dataset based on
multimodal corpus, Conceptual Captions \cite{sharma2018conceptual}, and multilingual corpus, Wikipedia. 
We evaluate \modelshort\ on multilingual image-text retrieval task on two datasets: Multi30K \cite{elliott2016multi30k, elliott2017findings} and MSCOCO \cite{chen2015microsoft,mcoco1,mcoco2}.
Panlex\footnote{ \href{https://panlex.org}{https://panlex.org}} is used as the bilingual dictionary during \textit{Multimodal Code-switched Training}.

\subsubsection{Pre-training Corpus}

\paragraph{Conceptual Captions}
We use Conceptual Captions \cite{sharma2018conceptual} as the multimodal corpus. It contains 3.3 million English image-caption pairs harvested from the Web and does not contain any non-English text.

\paragraph{Wikipedia}
We use sentences extracted from the Wikipedia dump as the multilingual corpus. It includes 101G sentences covering 100 languages without any vision information.

\subsubsection{Fine-tuning and Evaluation Corpus}

\paragraph{Multi30K}
This dataset extended Flickr30K \cite{young2014flick} from English (\texttt{en}) to German (\texttt{de}), French (\texttt{fr}) and Czech (\texttt{cs}). It contains 31,783 images and provides five captions per image in English and German and one caption per image in French and Czech. The train, dev, and test splits are defined in \cite{young2014flick}.

\paragraph{MSCOCO}
This dataset contains 123,287 images and provides five captions per image in English (\texttt{en}), but fewer in Chinese (\texttt{zh}) and Japanese (\texttt{ja}). STAIR Captions \cite{yoshikawa2017stair} extended MSCOCO \cite{chen2015microsoft} with 820K Japanese captions for COCO images. \cite{mcoco2} extended MSCOCO \cite{chen2015microsoft} with Chinese captions for 20K images. We use the same train, dev, and test splits for English and Japanese as defined in \cite{karpathy2015deep}. As for Chinese, we use the COCO-CN split \cite{mcoco2}.

\subsubsection{Code-switched Dictionary}

The word-level bilingual dictionaries used by Code-switched training are from Panlex, the world’s largest open-source lexical translation database. We extract top 50 scale English to other language bilingual dictionaries.

\subsection{Training Details}

\paragraph{Pre-training Details}
Similar to previous vision-language pre-trained models, the \modelshort\ model uses the same model architecture as BERT \cite{devlin2019bert}. We initialize \modelshort\ with XLM-R \cite{conneau2019unsupervised} and continue pre-training on our data. We use the same vocabulary as XLM-R \cite{conneau2019unsupervised}, which includes 250K BPE tokens and covers 100 languages.
We set the dropout rate to 0.1 and the max input length to 128. We use Adam Optimizer \cite{kingma2015adam} with a linear warm-up \cite{vaswani2017attention} and set the learning rate to $1 \times 10^{-4}$. The total batch size is 1,024 after gradient accumulation. The pre-training stage takes about seven days to converge on 8 V100 GPUs. We use \textit{Multimodal Code-switched Training} with all top 50 languages from Panlex.

\paragraph{Fine-tuning Details}
The batch size is set to 512, and we sample three negative cases for each positive case in \tasksixshort. We experiment with different numbers of negative samples in ${\{1,3,5\}}$, and find three yields the best results.
%To obtain the number of negative samples in image-text retrieval, we formulate it as a ranking problem. We respectively set 1,3 and 5 as the number of negative samples and achieve the best results on 3.
We use Adam Optimizer with $\beta_1=0.9$, $\beta_2=0.98$ and $5 \times 10^{-5}$ as the hyper-parameters of learning rate.

\subsection{Baselines}
We compare our work with several related work \cite{kim2019mule, wang2018learning,gella2017image,wehrmann2019language,burns2020learning}, which are trained on downstream task datasets (MSCOCO and Multi30K) directly without pre-training. In addition, to make the comparison as fair as possible, we take Unicoder-VL as another baseline, as it employs the same pre-training data during image-language pre-training. 

Among the baselines, SMALR \cite{burns2020learning} uses machine translation to augment Multi30K and MSCOCO. But considering that applying machine translation to translate English to all other supported languages lacks generalization and requires a large amount of translators, we leave this as an option for future work. Moreover, note that MULE is using different dev/test splits of MSCOCO compared with other models.

It is also worth noticing that word-level dictionaries are only used in M$^3$P, as the \textit{Multimodal Code-switched Training} is firstly used in multilingual multimodal pre-training.

\subsection{Evaluation Settings}

Multilingual image-text retrieval is the task of finding the most relevant images given input texts in different languages, or vice versa. We use mean Recall (mR) as our metric, which is an averaged score of Recall@1, Recall@5, and Recall@10 on image-to-text retrieval and text-to-image retrieval tasks. 

We compare \modelshort\ with baseline methods on multilingual image-text retrieval in four different settings:

(i) \textit{\textbf{w/o fine-tune}}: apply \modelshort\ to all test sets directly to obtain the evaluation results without fine-tuning. 

(ii) \textit{\textbf{w/ fine-tune on en}}: fine-tune \modelshort\ on English and then apply the fine-tuned model to all test sets.

(iii) \textit{\textbf{w/ fine-tune on each}}: fine-tune \modelshort\ on each language and apply each model to the test set of this language.

(iv) \textit{\textbf{w/ fine-tune on all}}: fine-tune \modelshort\ for all languages using the merged labeled data and then apply the fine-tuned model to all test sets.

%% file: 05_results.tex
\section{Results and Analysis}

In this section, we show the evaluation results of \modelshort\ compared with existing work and conduct ablation studies in order to better understand the effect of the model.

\subsection{Overall Results}

\begin{table*}[t]
\small
\centering
\begin{tabular}{l|cccc|ccc}
\toprule
\multirow{2}{*}{Model} & \multicolumn{4}{c|}{Multi30K} & \multicolumn{3}{c}{MSCOCO}\\ 
& \texttt{en} & \texttt{de} & \texttt{fr} & \texttt{cs} & \texttt{en} & \texttt{ja} & \texttt{zh} \\ 
\midrule
\midrule
\multicolumn{8}{l}{\textit{Monolingual supervised results}}\\
\midrule
EmbN~\cite{wang2018learning} & 72.0 & 60.3 & 54.8 & 46.3 & 76.8 & 73.2 & 73.5  \\
PAR. EmbN~\cite{gella2017image} & 69.0 & 62.6 & 60.6 & 54.1 & 78.3 & 76.0 & 74.8 \\ 
S-LIWE~\cite{wehrmann2019language} & 76.3 & 72.1 & 63.4 & 59.4 & 80.9 & 73.6 & 70.0  \\
MULE~\cite{kim2019mule} & 70.3 & 64.1 & 62.3 & 57.7 & 79.0 & 75.9 & 75.6  \\
SMALR~\cite{burns2020learning} & 74.5 & 69.8 & 65.9 & 64.8 & 81.5 & 77.5 & 76.7  \\
\midrule
\midrule
\multicolumn{8}{l}{\textit{Monolingual results with multimodal pre-training}}\\
\midrule
Unicoder-VL (w/o fine-tune)~\cite{li2019unicodervl} & 72.0 & - & - & - & 63.7 & - & -  \\
Unicoder-VL (w/ fine-tune on en)~\cite{li2019unicodervl} & \textbf{88.1} & - & - & - & \textbf{89.2} & - & -  \\
\midrule
\midrule
\multicolumn{8}{l}{\textit{Multilingual results with multimodal pre-training}}\\
\midrule
\modelshort\ (w/o fine-tune) & 57.9 & 36.8 & 27.1 & 20.4 & 63.1 & 33.3 & 32.3 \\
\modelshort\ (w/ fine-tune on en) & 87.4 & 58.5 & 46.0 & 36.8
 & 88.6 & 53.8 & 56.0 \\
\modelshort\ (w/ fine-tune on each) & 87.4 & 82.1 & 67.3 & 65.0 & 88.6 & 80.1 & 75.8 \\
\modelshort\ (w/ fine-tune on all) & 87.7 & \textbf{82.7} & \textbf{73.9} & \textbf{72.2} & 88.7 & \textbf{87.9} & \textbf{86.2} \\
\bottomrule
\end{tabular}
\caption{\label{retrieval-on-coco} Multilingual image-text retrieval results on Multi30K and MSCOCO. 
The metric is the mean Recall (mR). Each \textbf{bold number} indicates the best mR score in that column.
We report the mR results of Unicoder-VL on the English test set, as it is pre-trained based on the same image-caption corpus (i.e., Conceptual Captions) with \modelshort\ .
}
\end{table*}

From Table \ref{retrieval-on-coco}, we have several observations: (1) Our \modelshort\ model obtains the state-of-the-art results in all non-English languages, which shows its exciting multilingual multimodal transfer capability.
(2) Similar to the observations reported in Unicoder \cite{huang2019unicoder,liang2020xglue}, the two \textit{fully-supervised} settings (iii) \textit{w/ fine-tune on each} and (iv) \textit{w/ fine-tune on all} can lead to the best results. This means the same sentence in different languages may capture complementary information to help improve performance. (3) Comparing to Unicoder-VL that is pre-trained using English image-caption corpus (i.e. Conceptual Captions) only, \modelshort\ performs worse on the English test set. The possible reason could be that, \modelshort\ needs to balance its multilingual capability over 100+ languages, rather than on English only. (4) In both setting (i) \textit{w/o fine-tune} and setting (ii) \textit{w/ fine-tune on en}, integrating \textit{Multimodal Code-switched Training} (MCT) into M$^3$P can bring significant gains on non-English datasets, which demonstrates good multilingual transfer ability of \textit{Multimodal Code-switched Training} in the zero-shot setting. It is expected to see such gains become smaller in setting (iii) \textit{w/ fine-tune on each} and setting (iv) \textit{w/ fine-tune on all}, as M$^3$P can learn alignments between images and languages from labeled data directly.

\subsection{Ablation Studies}

\label{cs}
Although we achieve good results under different settings, we want to deep dive into more aspects of \modelshort: (1) whether \textit{Multimodal Code-switched Training} (MCT) can provide a positive effect under all settings; (2) whether the number of languages used in MCT affects the performance; (3) whether different pre-training tasks affect the performance.

\subsubsection{The Impact of MCT}

To verify whether the \textit{Multimodal Code-switched Training} (MCT) strategy can provide a positive effect in different settings, we compare the performance of \modelshort\ without MCT and \modelshort\ with MCT under all fine-tuning settings.

\label{code-int}
\begin{table}[!htp]
\small
\centering
\begin{tabular}{l|cccc}
\toprule
\multirow{2}{*}{Setting} & \multicolumn{4}{c}{Multi30K} \\
& \texttt{en} & \texttt{de} & \texttt{fr} & \texttt{cs}   \\
\midrule
\midrule
\multicolumn{5}{l}{\textit{w/o fine-tune}}\\
\midrule
\modelshort\ w/o MCT & 54.9 & 28.9 & 25.2 & 13.5  \\
\quad\ w/ MCT & \textbf{57.9} & \textbf{36.8} & \textbf{27.1} & \textbf{20.4} \\
\midrule
\midrule
\multicolumn{5}{l}{\textit{w/ fine-tune on en}}\\
\midrule
\modelshort\ w/o MCT & 86.0 & 48.6 & 37.1 & 34.6  \\
\quad\ w/ MCT & \textbf{87.4} & \textbf{58.5} & \textbf{46.0} & \textbf{36.8}  \\
\midrule
\midrule
\multicolumn{5}{l}{\textit{w/ fine-tune on each}}\\
\midrule
\modelshort\ w/o MCT & 86.0 & 80.2 & 67.1 & \textbf{66.2}  \\
\quad\ w/ MCT & \textbf{87.4} & \textbf{82.1} & \textbf{67.3} & 65.0 \\
\midrule
\midrule
\multicolumn{5}{l}{\textit{w/ fine-tune on all}}\\
\midrule
\modelshort\ w/o MCT & 86.7 & 82.0 & 73.5 & 70.2  \\
\quad\ w/ MCT & \textbf{87.7} & \textbf{82.7} & \textbf{73.9} & \textbf{72.2} \\
\bottomrule
\end{tabular}
\caption{\label{CSP-abMCT}The impact of MCT for multilingual image-text retrieval. The metric is the mean Recall (mR). Each \textbf{bold number} indicates the best mR score. }
\end{table}

For each setting in Table \ref{CSP-abMCT}, we observe: (1) MCT improves the performance on almost all languages, which shows its exciting robustness and expansibility, and (2) in both setting (i) and setting (ii), integrating MCT into M$^3$P can bring significant gains on non-English datasets, which demonstrates the good multilingual transferability of MCT. It is expected to see such gains become smaller in settings (iii) and (iv), as M$^3$P can learn alignments between images and languages from labeled data directly.

\subsubsection{The Impact of Number of Languages in MCT}

\begin{table}[!htp]
\small
\centering
\begin{tabular}{l|cccc}
\toprule
\multirow{2}{*}{Setting} & \multicolumn{4}{c}{Multi30K} \\
& \texttt{en} & \texttt{de} & \texttt{fr} & \texttt{cs}   \\
\midrule
\midrule
\modelshort\ w/o MCT & 54.9 & 28.9 & 25.2 & 13.5  \\
\quad\ w/ 3 languages MCT & 56.4 & 37.1 & 28.7 & 23.0 \\
\quad\ w/ 5 languages MCT & 58.2 & 36.7 & 26.9 & 23.6 \\
\quad\ w/ 50 languages MCT (Full) & 57.9 & 36.8 & 27.1 & 20.4 \\
\bottomrule
\end{tabular}
\caption{\label{CSP-number}Impact of number of languages in \textit{Multimodal Code-switched Training} (MCT). The metric is the mean Recall (mR). "Full" represents the model pre-trained with all Code-switching languages.}
\end{table}

To verify whether the number of languages influences the performance of \textit{Multimodal Code-switched Training} (MCT), we conduct an experiment by pre-training \modelshort\ by MCT with different numbers of languages and evaluate the model directly without fine-tuning.
We pre-train \modelshort\ with the following settings: pre-train \modelshort\ without MCT, pre-train \modelshort\ with MCT on 3 languages (\texttt{de}, \texttt{fr}, \texttt{cs}), 5 languages (\texttt{de}, \texttt{fr}, \texttt{cs}, \texttt{ja}, \texttt{zh}), and all 50 languages.

In Table \ref{CSP-number}, we can find that, for languages like \texttt{de} and \texttt{fr}, there is no significant difference under different settings. On the contrary, for languages like \texttt{en} and \texttt{cs}, \modelshort\ achieves the best performance when MCT is activated with 5 languages.
%and \texttt{cs} drops 3.2 in the full model. 
This implies that activating MCT on more languages can lead to more noise due to a higher probability of inaccurate translation. This noise may improve the robustness of the model but make the model harder to be well-trained.

\subsubsection{The Impact of Proposed Tasks}

We want to find whether each component during pre-training positively affects the performance and try to explain how they gain the performance by conducting several ablation experiments. Since \textit{Multimodal Code-switched Training} (MCT) influences each task's target, we conduct the ablation experiments on \modelshort\ without MCT and fine-tune each model on the dataset of each language to compare the 
performance.

\begin{table}[t]
\small
\centering
\begin{tabular}{l|cccc}
\toprule
\multirow{2}{*}{Setting} & \multicolumn{4}{c}{Multi30K}  \\
& \texttt{en} & \texttt{de} & \texttt{fr} & \texttt{cs}  \\
\midrule
\midrule
\modelshort\ & \textbf{86.0} &  \textbf{80.2} & \textbf{67.1} & \textbf{66.2}  \\
%\modelshort\ & \textbf{87.4} &  \textbf{82.1} & \textbf{67.3} & \textbf{65.0}  \\
\quad\ w/o xMLM   & 79.6 & 70.8 & 56.4 & 54.3   \\
\quad\ w/o MC-MLM    & 84.3 & 76.2 & 64.1 & 62.2     \\
\quad\ w/o MC-MRM    & 85.5 & 77.9 & 65.0 & 63.9     \\
\quad\ w/o MC-VLM    & 75.4 & 68.3 & 52.7 & 50.9    \\
\bottomrule
\end{tabular}
\caption{\label{ablation-Multi30K}Ablation study on multilingual image-text retrieval. The metric is the mean Recall (mR). Each \textbf{bold number} indicates the best mR score in that column. }
\end{table}

As shown in Table \ref{ablation-Multi30K}, we can observe that: (1) MC-VLM provides the most considerable improvement (+10.6 on en) to the model among all four sub-tasks during the pre-training stage. We suggest this is because the MC-VLM sub-task successfully models the relationship between image and text. (2) xMLM shows a great impact on non-English results compared with English results, which shows that xMLM will improve the capability of multilinguality. (3) MC-MLM and MC-MRM also show good support to the results in all languages, which we suggest these two tasks will help the model learn the 
knowledge of multimodality. (4) When combining all tasks, we obtain the highest gain in all languages.

\subsection{Expanding MCT to Fine-tuning}

\begin{table}[!htp]
\small
\centering
\begin{tabular}{l|cccc}
\toprule
\multirow{2}{*}{Setting} & \multicolumn{4}{c}{Multi30K} \\
& \texttt{en} & \texttt{de} & \texttt{fr} & \texttt{cs}   \\
\midrule
\midrule
\multicolumn{5}{l}{\textit{Pre-trained without MCT}}\\
\midrule
\modelshort\ (w/ Normal Fine-tune) & \textbf{86.0} & 48.6 & 37.1 & 34.6 \\
\modelshort\ (w/ MCT Fine-tune)  & 85.4 & \textbf{67.8} & \textbf{59.2} & \textbf{54.0} \\
\midrule
\midrule
\multicolumn{5}{l}{\textit{Pre-trained with MCT}}\\
\midrule
\modelshort\ (w/ Normal Fine-tune) & \textbf{87.4} & 58.5 & 46.0 & 36.8  \\
\modelshort\ (w/ MCT Fine-tune)  & 86.4 & \textbf{71.8} & \textbf{62.3} & \textbf{59.6} \\
\bottomrule
\end{tabular}
\caption{\label{CSP-ft}The results of expanding MCT to fine-tuning for multilingual image-text retrieval. The metric is the mean Recall (mR). Each \textbf{bold number} indicates the best mR score under the setting. \textit{Normal Fine-tune} represents fine-tuning with English data directly and \textit{MCT Fine-tune} represents fine-tuning with Code-switched English data.}
\end{table}

Similar to MC-VLM, we use Code-switched data to fine-tune \modelshort\ on Multi30K. The results in Table \ref{CSP-ft} show: (1) \textit{Multimodal Code-switched Training} (MCT) can bring a large margin for non-English language probably because of the lack of labeled image-non English caption pairs during the pre-training stage or fine-tuning stage. (2) Employing MCT into the fine-tuning stage for the model, whatever pre-trained by, will achieve a large increase in non-English performance.
(3) MCT in fine-tuning is more effective than MCT in pre-training, which may be explained by that the model can learn multilinguality in a more specific task.
(4) The best results can be achieved when MCT replaces English in both the pre-training and fine-tuning stages.

\subsection{Qualitative Studies on MCT}

To further explore how \textit{Multimodal Code-switched Training}  (MCT) affects the model, we randomly select some text-image pairs generated from it. We want to figure out why \textit{Multimodal Code-switched Training} is very effective on non-English languages and whether it has any limitations.

\begin{figure} [!htp]
    \centering
    \includegraphics[width=8cm]{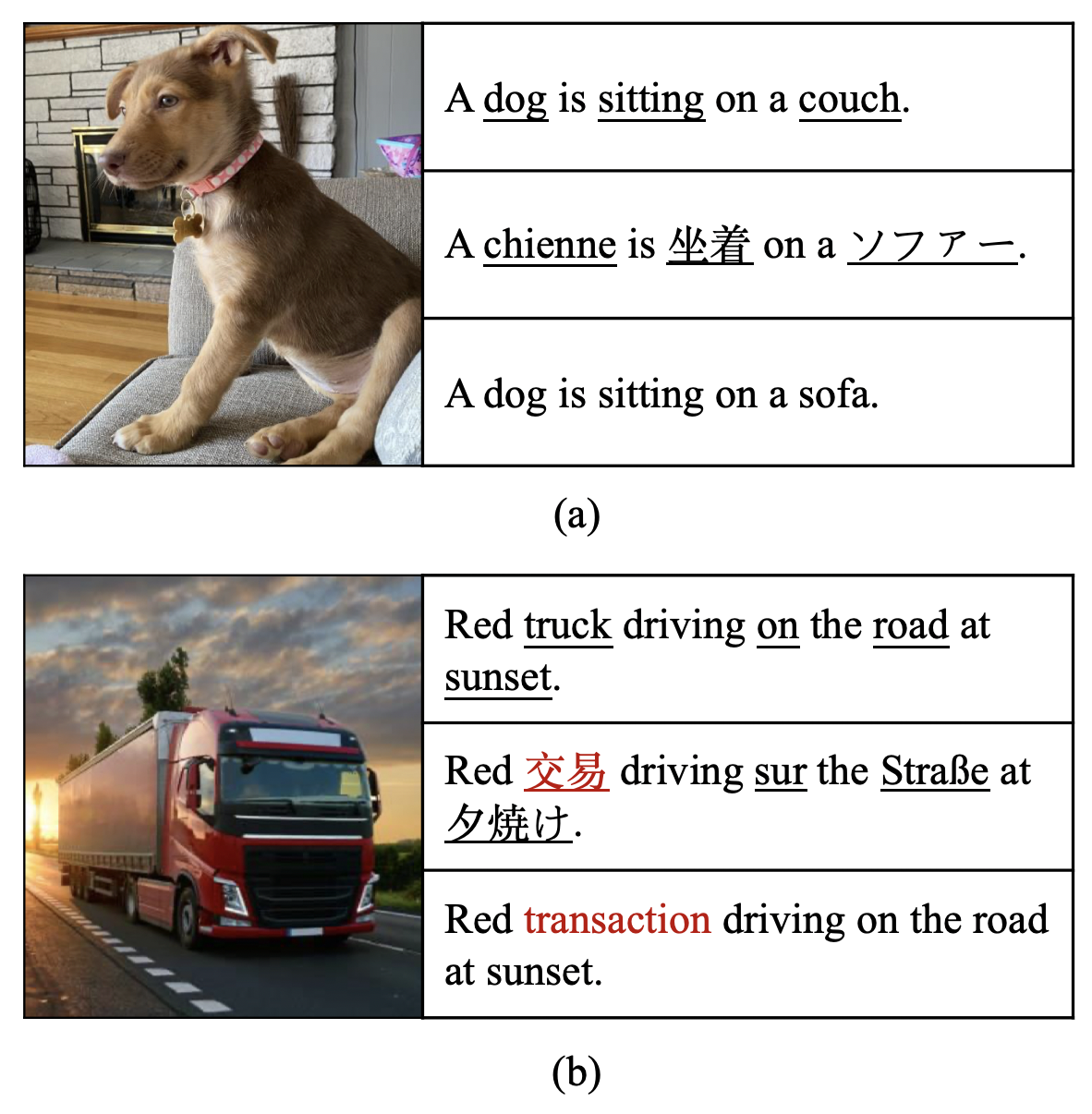}
    \caption{Qualitative study for \textit{Multimodal Code-switched Training} (MCT). The first row in each table is the original text, and the second row in each table is the Code-switched text. We add the meaning of the Code-switched text in English in the third row of each table.}
    \label{case}
\end{figure}

As Figure \ref{case} (a) shows, the meaning of the code-switched text generated by \textit{Multimodal Code-switched Training} (MCT) is almost the same as that of the original text. Although there are some small differences between the original text (first row) and the generated text translated back to English (third row), it has no influence on the training quality, which demonstrates the reason why MCT brings gains. The key idea of using MCT in M$^3$P is to let the model see more Code-switched text and image pairs and learn the joint multilingual multimodal representations from such pairs directly. We guess this helps the model learn richer information of each token from the multilingual context.

We did not consider the grammar or syntax correctness of the Code-switched sentences generated by replacing words in the English sentences with their word translations in other languages. The pre-trained models can learn such knowledge from well-formed multilingual sentences and English caption sentences. Since we don't have image-caption pairs or high-quality machine translation engines to generate such data for most languages, generating Code-switched sentences is the most effective way to let M$^3$P directly see more alignments between non-English languages and images.

Hence, because of the high accuracy of translation from MCT, multilingual results will significantly increase when no non-English multimodal data is available. However, when the model can access high-quality multilingual multimodal data, the noise from MCT may limit its performance. In Figure \ref{case} (b), we show a negative case in Code-switched text. MCT faultily changes the meaning of the original text. We leave this as future work to solve this problem.

%% file: 06_conclusion.tex
\section{Conclusion}
We have presented in this paper a new pre-trained model \modelshort\ which combines Multilingual Pre-training and Multimodal Pre-training into a unified framework via Multitask Pre-training for multilingual multimodal scenarios. We proposed Multimodal Code-switched Training to further alleviate the issue of lacking enough labeled data for non-English multimodal tasks and avoid the tendency to model the relationship between vision and English text.